\documentclass[10pt,twocolumn,letterpaper]{article}

\usepackage{cvpr}      

\usepackage{graphicx}
\usepackage{amsmath}
\usepackage{amssymb}
\usepackage{booktabs}
\usepackage{algpseudocode}
\usepackage{algorithm}
\usepackage[accsupp]{axessibility}
\algnewcommand{\algorithmicor}{\textbf{ or }}
\algnewcommand{\algorithmicsort}{\textbf{sort }}
\algnewcommand{\algorithmicmap}{\textbf{project }}
\algnewcommand{\OR}{\algorithmicor}
\algnewcommand{\SORT}{\algorithmicsort}
\algnewcommand{\PROJECT}{\algorithmicmap}
\newcommand{\pluseq}{\mathrel{+}=}
\usepackage{multirow}
\usepackage{animate}

\newcommand*{\affaddr}[1]{#1}
\newcommand*{\affmark}[1][*]{\textsuperscript{#1}}
\newcommand*{\email}[1]{\texttt{#1}}

\usepackage[pagebackref,breaklinks,colorlinks]{hyperref}

\usepackage[capitalize]{cleveref}
\crefname{section}{Sec.}{Secs.}
\Crefname{section}{Section}{Sections}
\Crefname{table}{Table}{Tables}
\crefname{table}{Tab.}{Tabs.}

\def\cvprPaperID{6276} 
\def\confName{CVPR}
\def\confYear{2022}

\begin{document}

\title{XYLayoutLM: Towards Layout-Aware Multimodal Networks For Visually-Rich Document Understanding}

\author{%
Zhangxuan Gu\affmark[1,2], Changhua Meng\affmark[2], Ke Wang\affmark[2], Jun Lan\affmark[2], Weiqiang Wang\affmark[2], Ming Gu\affmark[2], Liqing Zhang\affmark[1]\thanks{Corresponding author.}\\
\affaddr{\affmark[1]{{MoE Key Lab of Artificial Intelligence,} Shanghai Jiao Tong University}},
\affaddr{\affmark[2]Ant Group}\\
\email{zhangxgu@126.com}\\
\email{\{changhua.mch,kaywang.wk,yelan.lj,weiqiang.wwq,guming.mg\}@antgroup.com}\\
\email{zhang-lq@cs.sjtu.edu.cn}
}

\maketitle

\begin{abstract}
Recently, various multimodal networks for Visually-Rich Document Understanding(VRDU) have been proposed, showing the promotion of transformers by integrating visual and layout information with the text embeddings. However, most existing approaches utilize the position embeddings to incorporate the sequence information, neglecting the noisy improper reading order obtained by OCR tools. In this paper, we propose a robust layout-aware multimodal network named XYLayoutLM to capture and leverage rich layout information from proper reading orders produced by our Augmented XY Cut. Moreover, a Dilated Conditional Position Encoding module is proposed to deal with the input sequence of variable lengths, and it additionally extracts local layout information from both textual and visual modalities while generating position embeddings. Experiment results show that our XYLayoutLM achieves competitive results on document understanding tasks.
\end{abstract}

\section{Introduction}\label{intro}

While significant progress has been made in natural language processing and visual understanding~\cite{devlin2018bert,chu2021conditional,dosovitskiy2020image,liu2021swin}, less attention has been paid to their challenging variant in the multimodal document understanding domain. The Visually-Rich Document Understanding (VRDU)~\cite{xu2020layoutlm} task requires combining the abundant image, text, and layout information from scanned/digital-born documents (images, PDFs, \emph{etc.}) Such technology can benefit a great variety of scenarios such as report/receipt understanding, automatical form filling, and document relation extraction. As a result, it is in great need of effective and efficient document understanding approaches.

To this end, researchers have developed sophisticated pipelines for tackling this task~\cite{xu2020layoutlm,xu2020layoutlmv2,xu2021layoutxlm,garncarek2020lambert,appalaraju2021docformer,li2021selfdoc,li2021structurallm}. Generally speaking, early attempts can be divided into the categories of textual-based~\cite{conneau2019unsupervised,chi2020infoxlm,garncarek2020lambert}, convolution-based~\cite{hao2016table,schreiber2017deepdesrt,katti2018chargrid,soto2019visual,zhong2019publaynet} and GCN-based~\cite{liu2019graph} methods. Text-based methods, \emph{e.g.}, XLM-RoBERT~\cite{conneau2019unsupervised} and InfoXLM~\cite{chi2020infoxlm}, usually rely on the representation ability of self-supervised models like Bert~\cite{devlin2018bert} pretrained on large datasets. Convolution-based method Chargrid~\cite{katti2018chargrid} utilized a fully convolutional network that predicted a segmentation mask and bounding boxes for document representation. More recently, \cite{liu2019graph} introduces a Graph Convolutional Networks based model to fuse the textual and visual feature from scanned documents. 

Although attempts like LayoutLM~\cite{xu2020layoutlm},  LayoutLMv2~\cite{xu2020layoutlmv2} and LayoutXLM~\cite{xu2021layoutxlm} have been made to tackle document understanding in a multimodal manner, they still confront two limitations: (1) They rely on the tokens and boxes from OCR~\cite{1998Gradient} tools without exploring the effect of reading orders. The proper reading orders refer to the well-organized readable token sequences, which may not be unique. Intuitively, the reading order of input tokens is crucial to many tasks such as language translation~\cite{vaswani2017attention} and VQA~\cite{2019Deep}. For example, the meaning of a sentence may be changed when we shuffle the words, resulting in mistakes during language translation. A common solution is to use position embeddings to denote such sequential order of input tokens. However, we find that multimodal models with widely-used relative position embeddings still suffer improper reading order. Proper reading orders implicitly include the layout information, which is essentially needed in VRDU tasks. (2) They usually leverage fixed-length absolute/relative position embeddings in transformers. Once the model is trained, it can not deal with the test data with longer token sequences. Although bilinear interpolation on position embeddings can be applied to the longer sequence, the performance is not satisfying. Recently, Conditional Position Encoding (CPE)~\cite{chu2021conditional} is proposed to deal with inputs of variable lengths in image classification tasks. It reshaped the input tokens to 2D features and dynamically extracted local neighbor context from input tokens with convolutions. However, since CPE is designed for only visual tokens, it can not handle 1D textual tokens in VRDU tasks.

\begin{figure}[t]
\centering
\includegraphics[width=0.46\textwidth]{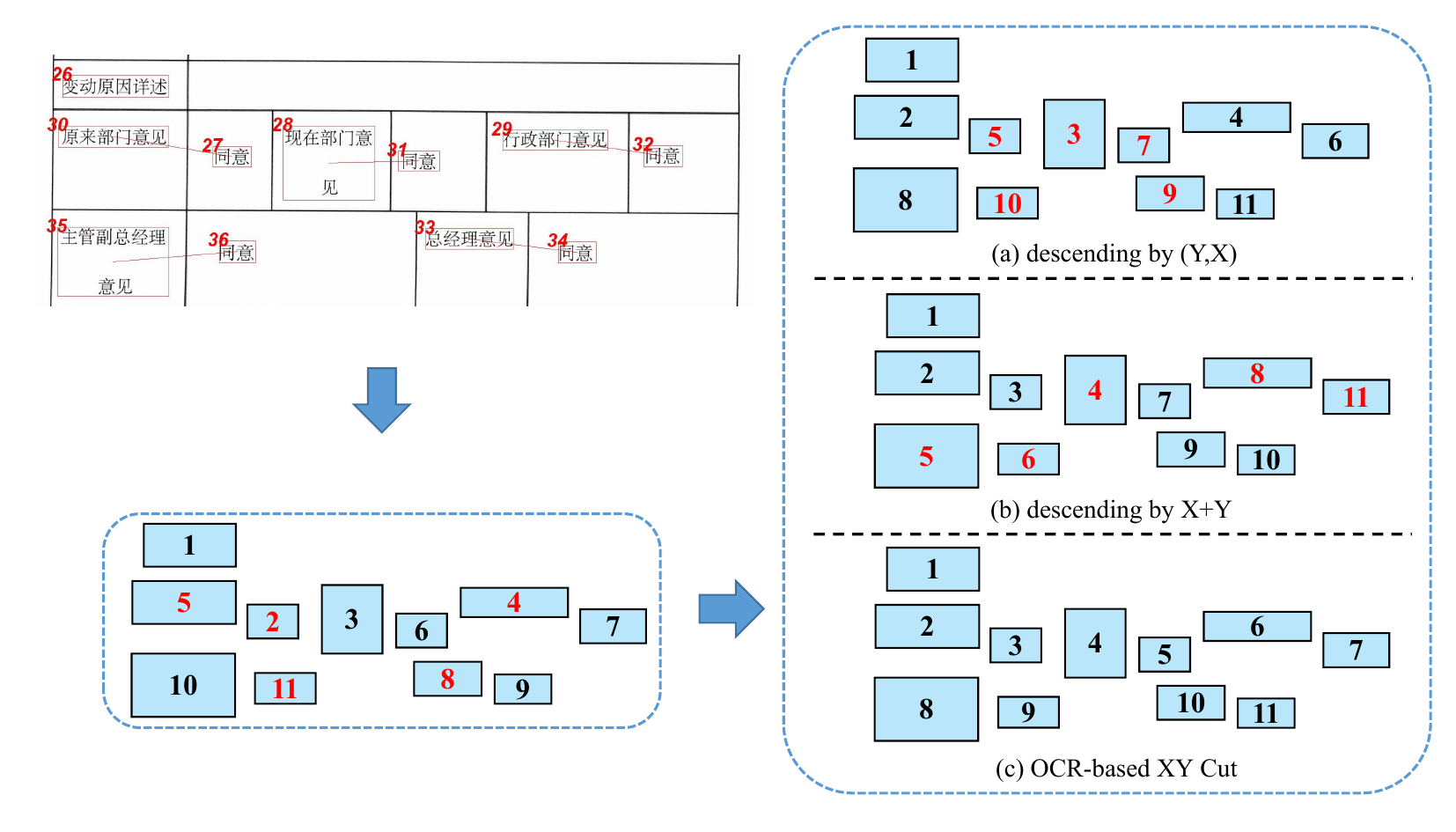}
\caption{An example from XFUN dataset. The reading order is denoted by the indices in the boxes while the red ones mean their orders are improper.}
\label{fig1}
\end{figure}

In this paper, we propose an improved version of LayoutLMv2~\cite{xu2020layoutlmv2}, \textbf{XYLayoutLM}. Instead of pretraining on large private/public document understanding datasets, XYLayoutLM focuses on the generation of position embeddings with two under-explored limitations in VRDU, \emph{i.e.}, improper reading orders, and the disability of dealing with longer sequence, as mentioned above.

Although it seems a fundamental requirement of multimodal tasks to have proper reading orders, it is non-trivial to directly obtain such reading orders from documents due to various formats, \emph{e.g.}, tables, and columns. Specifically, we show a form from XFUN~\cite{xu2021layoutxlm} dataset in Figure~\ref{fig1}. The default reading order is noisy(also in Figure~\ref{fig3}). Based on the boxes obtained by OCR tools, traditional sorting approaches such as arranging the tokens in a top-to-bottom and left-to-right way are not satisfying. For example, we list two simple heuristic rules in this figure, namely (a) descending first by Y-axis then X-axis, (b) descending by Y$+$X conditioned on the left-top points of the token boxes. However, the red indices in Figure~\ref{fig1} still highlight the tokens with improper reading orders. Finally, we utilize the XY Cut~\cite{ha1995recursive} (c) and successfully obtain one proper reading order.
Interestingly, some tokens in the same row may have different locations due to the noise in OCR recognition. It fails two heuristic rules which need the accuracy location of tokens. However, we can still obtain a series of proper reading orders for training by our proposed Augmented XY Cut as an augmentation strategy.

For input sequences of variable lengths, we utilize a novel Dilated Conditional Position Encoding (DCPE) module to adaptively generate position embeddings according to their input lengths with the dilated convolutions for extracting local layouts. We demonstrate that the XYLayoutLM can lead to better performance than previous LayoutLMs~\cite{xu2020layoutlm,xu2020layoutlmv2,xu2021layoutxlm}, which will benefit a great variety of real-world document understanding applications. We summarize our contributions as follows.

\begin{itemize}
    \item For the first time, Augmented XY Cut is proposed and utilized to sort the input tokens for generating different proper reading orders in VRDU tasks. It extracts and leverages the layout information to achieve competitive performances.
    \item To deal with input sequences of variable lengths, we propose a Dilated Conditional Position Encoding as the position embedding generator to adaptively process the 1D textual and 2D visual tokens. Benefitting from proper reading orders, DCPE can further extract rich local layouts of input tokens with dilated convolutions.
    \item Comprehensive experiments are conducted on VRDU datasets. Our XYLayoutLM achieves competitive performance among all listed VRDU approaches on semantic entity recognition and relation extraction tasks.
\end{itemize}

\section{Related Works}

\subsection{Visually-Rich Document Understanding}
Recently, transformers-based methods have been proved to be effective on many computer vision~\cite{chu2021conditional,dosovitskiy2020image,liu2021swin} and natural language process~\cite{chi2020infoxlm,devlin2018bert,garncarek2020lambert} domains. Among them, \cite{xu2020layoutlm,xu2020layoutlmv2,xu2021layoutxlm} proposed a series of transformer-based models focusing on VRDU tasks. As our baseline, LayoutXLM\cite{xu2020layoutlm} is the multilingual version of LayoutLMv2\cite{xu2020layoutlmv2}. They achieved impressive results by successfully combining textual, layout, and visual features. However, those methods may feed the input tokens to the transformer in the improper reading order caused by OCR tools on complex documents, which will harm VRDU performance. In this paper, we pay more attention to the under-explored challenge, \emph{i.e.}, the proper reading order of input tokens, which is significant to the model performance.

\subsection{Positional Encoding}
Positional encodings are commonly employed to incorporate the order of sequences because self-attention is permutation-equivalent. Existing research can be grouped into two categories: absolute and relative position encodings. When the transformer-based model was first proposed by \cite{vaswani2017attention}, they designed a delicate sin-cos function as the absolute positional encoding. After that, \cite{devlin2018bert} used a learnable absolute embedding which is an embedding of the same length to the input sequence. It can be jointly updated with the network weights during training. Recently, by considering the distance between tokens, \cite{shaw2018self} proposed to change position embedding from absolute way into the relative way. However, they can not handle the longer sequences with fixed-length position encodings. To this end, Conditional Position Encoding (CPE)~\cite{chu2021conditional} was proposed to deal with the input sequence of variable lengths in the image classification task. It generates position embeddings conditioned on the local context extracted by 2D convolutional layers. However, CPE can not be used in multimodal networks due to the 1D text features in document understanding tasks.

\begin{figure*}[h]
\centering
\animategraphics[width=0.95\textwidth, autoplay, loop]{2}{fig6/fig-}{1}{6}
\caption{The overview of \textbf{XYLayoutLM}. Different from LayoutXLM, our XYLayoutLM proposes Augmented XY Cut and DCPE to extract and utilize layout information for multimodal document understanding. Best viewed in Adobe Acrobat DC.}
\label{overview}
\end{figure*}




\subsection{Reading Order Detection}
Reading order detection~\cite{2003Bidimensional,2007A,learningtoorder,malerba2008machine,ferilli2014abstract,li2020end} aims to capture proper reading orders for documents. Generally speaking, humans tend to read documents left-to-right and up-to-bottom ways. However, such simple sorting rules may fail due to the tokens extracted by OCR tools on complex documents. Recently, \cite{wang2021layoutreader} proposed a multimodal network for reading order detection with a large benchmark made by tremendous complex documents. However, compared to our method, the labor for collecting 500k standard Word files and the time for training a LayoutReader\cite{wang2021layoutreader} cannot be ignored.
Meanwhile, the inference time for LayoutReader on reading order detection is much longer than our method(see Appendix). 
In this paper, we proposed a simple yet effective augmentation algorithm based on XY Cut~\cite{ha1995recursive} to obtain different proper reading orders. 



\section{Methodology}

\subsection{Overview}

The overall XYlayoutLM architecture is depicted in Figure~\ref{overview}. The model takes the images, textual tokens, and text locations (boxes) as the input. Visual tokens are obtained by adaptively pooling the feature map of ResNeXt-101 to $7\times 7$. Then we flatten and concentrate it with the textual tokens to form the input token sequence following~\cite{xu2021layoutxlm}. Two individual position embedding generators are utilized for encoding the tokens into position embeddings and box embeddings. Different from the baseline model LayoutXLM~\cite{xu2021layoutxlm}, our XYLayoutLM has two advantages: (1) An Augmented XY Cut module is proposed to sort the input tokens for different proper reading orders, which will contribute to the model performances by leveraging the layout information in reading order. (2) Instead of generating position embeddings with fixed-length Multilayer Perceptron (MLP), we propose the Dilated Conditional Position Encoding (DCPE) module to handle the input tokens of variable lengths from texts and images. 
In this section, we will first briefly introduce the baseline LayoutXLM~\cite{xu2021layoutxlm} and then elaborate on the above-proposed components.


\subsection{Review of LayoutXLM}
Recall that LayoutXLM~\cite{xu2021layoutxlm} accepts inputs of three modalities: text, image, and layout (\emph{i.e.}, token locations). The input of each modality is converted to an embedding sequence by a fixed-length MLP operated on the position indices as shown in Figure~\ref{overview}. The text and image embeddings are concatenated, plus the layout embedding to get the input embedding. After that, the input embeddings are encoded by a transformer with the spatial-aware self-attention mechanism within and between modalities. Finally, the visual/text token representations outputted by the transformer are used in the document understanding tasks. Since the architecture of self-attention layers is not our main concern, we omit it here and refer readers to \cite{xu2020layoutlmv2,xu2021layoutxlm} for the details.

\subsection{Proper Reading Orders}\label{xycut}

How to obtain proper reading orders of documents like forms and receipts is an open question. Intuitively, it is possible to infer how token boxes are aligned and where significant horizontal and vertical gaps are present from projection profiles. Hence, the projection profiles of token boxes can be used to determine the reading order. In this section, we first introduce the projection profiles of token boxes and then present the Augmented XY Cut algorithm.




\noindent \textbf{Projection Profiles.} 
Suppose we are given a set of token boxes $B=\{b_i\}_{i=1}^K$, where each $b_i = [x_1^i,y_1^i,x_2^i,y_2^i]\in \mathbb{Z}^4$ denotes a box and $K$ is the number of OCR extracted tokens. We also define the minimum and maximum token locations in $B$ as $(x_{min},y_{min})$ and $(x_{max},y_{max})$. Then the horizontal mapping $H_{b_i}$ of box $b_i$ is formulated as a indicative function:
\begin{equation}
H_{b_i}(y) = \left\{
\begin{aligned}
1 & , &y_1^i \le y \le y_2^i \\
0 & , & otherwise
\end{aligned}
\right.
\end{equation}
where $y\in \mathbb{Z}^{[y_{min},y_{max}]}$. For a location $y$ on Y-axis, $H_{b_i}(y)$ effectively mean whether $y$ is in the projection interval $[y_1^i, y_2^i]$. Based on $H_{b_i}$, we can define the horizontal projection profile of the set $B$ by summing all horizontal mapping functions of individual boxes:

\begin{equation}
H_B(y)=\sum_{i=1}^KH_{b_i}(y).
\end{equation}
The values of $H_B(y)$ represent how many token boxes are projected onto the Y-axis that covers the input variable $y$.

Similar to $H_B(y)$, the vertical projection profile of $B$ can be denoted as follows:
\begin{equation}
V_B(x)=\sum_{i=1}^KV_{b_i}(x).
\end{equation}
where 
\begin{equation}
V_{b_i}(x) = \left\{
\begin{aligned}
1 & , &x_1^i \le x \le x_2^i \\
0 & , & otherwise
\end{aligned}
\right.
\end{equation}
is the vertical mapping on $b_i$ with $x\in \mathbb{Z}^{[x_{min},x_{max}]}$.

\noindent \textbf{Valleys In Projection Profiles.} For simplicity, let us take the horizontal projection profile $H_B$ as an example. As we mentioned before, after we have projected the token boxes $B$ to Y-axis to get corresponding intervals $\{[y_1^i, y_2^i]\}_{i=1}^K$, $H_B(y)$ is like a histogram for counting the intervals that cover $y$. As a result, there might be some valleys in the histogram. The valley here is defined as $y^*\in \mathbb{Z}^{[y_{min},y_{max}]}$ that meet the condition $H_B(y^*)=0$. There is no token box in the valleys. Thus, valleys of projection profiles can determine where the division has to take place.

\noindent \textbf{Augmented XY Cut Algorithm.} 
Traditional XY Cut is a heuristic divide and conquer algorithm first proposed by \cite{nagy1984hierarchical} to segment a sentence into words according to the values of every pixel. In 1995, \cite{ha1995recursive} utilized it to decompose documents (like newspapers) by applying XY Cut on the connected components. However, no current work explores the XY Cut in multimodal models or other deep learning transformers for obtaining proper reading orders.

As mentioned in the introduction, the token locations are often recognized with noise. Instead of expensive human annotations of reading orders, we propose an augmentation strategy on these noisy locations to generate different proper reading orders during network training. We believe the proper reading orders implicitly contain important layout information, significantly benefiting document understanding tasks. We only perform traditional XY Cut as a pre-process during the inference stage.

With the projection profiles and valleys defined above, Augmented XY Cut can be explained as follows. To better introduce it, we construct an XY Tree for recording the reading order while performing Augmented XY Cut. As shown in Algorithm~\ref{alg:xycut} and Figure~\ref{overview}(b), we take the boxes set $B$ as input, and the algorithm will output an index array $O=\{s(i)\}_{i=1}^K$ as one proper reading order. At first, we create a root node without any index. Then, unlike traditional XY Cut, we propose an augmentation strategy based on three thresholds: $\lambda_x,\lambda_y,\theta$. Specifically, $\lambda_x(\lambda_y)$ are thresholds that determine whether we shift a box on X-axis (Y-axis). If so, we will shift a box with $\theta\cdot v_x(\theta\cdot v_y)$ pixels in one direction. For every box, we generate two random values $v_x,v_y$ from $N(-1,1)$. If $|v_x|>\lambda_x(|v_y|>\lambda_y)$, then this box will be shifted with $\theta\cdot v_x(\theta\cdot v_y)$ pixels with the direction according to the sign of $v_x(v_y)$. In this paper, we set hyper-parameters as $0.5,0.5,5$ by default according to the experimental results.

At each step, the projection profiles are calculated in either horizontal or vertical directions. Then a division is performed at the valleys in the corresponding projection profile to obtain several clusters. Their token boxes are gathered as the new child nodes of XY Tree in descending order. To obtain a proper reading order of length $K$ is now divided into sub-tasks in each cluster with the number of boxes as the sequence length. The process is repeated recursively until no sufficient valleys are left in both profiles. If a cluster has more than one box and it can not be divided by both horizontal and vertical projection profiles, then the reading order inside this cluster will follow the heuristic rules, \emph{e.g.}, descending first by Y-axis then X-axis. Finally, the output reading order is obtained by collecting the indices on leaves of the XY Tree \emph{w.r.t} the tree height.

\begin{algorithm}[h]
\caption{Augmented XY Cut Algorithm}\label{alg:xycut}
\begin{algorithmic}[1]
\Require boxes: $B=\{b_i\}_{i=1}^K$, thresholds: $\lambda_x,\lambda_y,\theta$
\Ensure proper reading order: $O = \{s(i)\}_{i=1}^K$
\State Create a root node.     \Comment{Init XY Tree}
\State Do augmentation on $B$ with $\lambda_x,\lambda_y,\theta$.     \Comment{Augmentation}
\State Find the valleys of horizontal ($H_B$) or vertical ($V_B$) projection profiles.
\State Do divisions at valleys. Whenever divisions are made, create a new child node. At each recursion level, horizontal and vertical divisions alternate.
\State Do Step 3-4 recursively until no further divisions are possible.
\State Gather the indices on leaves as output $O$.
\end{algorithmic}
\end{algorithm}

To better explain the generation of XY Tree, we take Figure~\ref{overview}(b) as an example. In the first step, we horizontally project all seven boxes to the Y-axis by calculating the values of horizontal projection profile $H_B(y)$. We find there is only one valley and then perform a division to get two clusters in descending order according to their location on Y-axis (see $1$ and $2\cup3\cup4\cup5\cup6\cup7$ in Figure~\ref{overview}(b)). 
The first cluster only has one element, \emph{i.e.} box $1$, and thus it is the first leaf of XY Tree. The second cluster has six elements with the candidate order array $2\cup3\cup4\cup5\cup6\cup7$ and thus is fed to the second step with vertical projection profiles. In the second step, two valleys are detected, and thus, the order of $2$ and $7$ are decided as the leaves of XY Tree while $3\cup4\cup5\cup6$ still need further divisions. By iteratively performing horizontal and vertical projections, we can obtain the final reading orders on the tree leaves. The pseudo-code is shown in the Appendix.

\subsection{Dilated Conditional Position Encoding}\label{dcpe}

Conditional Position Encoding (CPE)~\cite{chu2021conditional} aims to generate a various-length position embedding for different inputs in image classification task. Specifically, it reshapes the flattened input sequence $X$ back to $X'$ in the 2D visual space. Then, convolutional layers are repeatedly applied to the $X'$ to produce the positional embedding $E$ with proper kernel and padding size to keep the resolution. Finally, the position embedding $E$ is flattened and added to the token embeddings as the transformer input.

\begin{table*}[h]
\centering
\begin{tabular}{c|c|c|ccccccc}
\toprule[1.5pt]
Task                  & Methods                             & XFUN Avg.                              & ZH                                     & JA                                     & ES                                     & FR                                     & IT                                     & DE                                     & PT                                     \\ \hline \hline
                      & XLM-RoBERT\cite{conneau2019unsupervised}                                   & 0.7047                                 & 0.8774                                 & 0.7761                                 & 0.6105                                 &  0.6743 & 0.6687                                 & 0.6814                                 & 0.6818                                 \\
                      & InfoXLM\cite{chi2020infoxlm}                                    & 0.7207                                 & 0.8868                                 & 0.7865                                 & 0.6230                                 & 0.7015                                 & 0.6751                                 & 0.7063                                 & 0.7008                                 \\
                      & LayoutXLM\cite{xu2021layoutxlm}                                   & 0.8056                                 & 0.8924                                 & 0.7921                                 & 0.7550                                 & 0.7902                                 & 0.8082                                 & 0.8222                                 & 0.7903                                 \\
                      & LayoutXLM+CPE                                     &  0.8047                                      & 0.8776                                 & 0.7909                                 & 0.7551                                 & 0.7908                                     & 0.8063                                       & 0.8227                                       &  0.7898                                      \\
\multirow{-5}{*}{SER} & XYLayoutLM     & {\textbf{0.8204}} & {\textbf{0.9176}} & {\textbf{0.8057}} & {\textbf{0.7687}} & {\textbf{0.7997}} & {\textbf{0.8175}} & {\textbf{0.8335}} & {\textbf{0.8001}} \\  \hline
                      & XLM-RoBERT\cite{conneau2019unsupervised}                                   & 0.4769                                 & 0.5105                                 & 0.5800                                 & 0.5295                                 & 0.4965                                 & 0.5305                                 & 0.5041                                 & 0.3982                                 \\
                      & InfoXLM\cite{chi2020infoxlm}                                      & 0.4910                                 & 0.5214                                 & 0.6000                                 & 0.5516                                 & 0.4913                                 & 0.5281                                 & 0.5262                                 & 0.4170                                 \\
                      & LayoutXLM\cite{xu2021layoutxlm}                                     & 0.6432                                 & 0.7073                                 & 0.6963                                 & 0.6896                                 & 0.6353                                 & 0.6415                                 & 0.6551                                 & 0.5718                                 \\
                      & LayoutXLM+CPE                                       &   0.6399                                     & 0.7059                                 & 0.6968                                 & 0.6812                                 &  0.6238                                      & 0.6399                                       &   0.6474                                     & 0.5723                                       \\
\multirow{-5}{*}{RE}  & XYLayoutLM     & {\textbf{0.6779}} & {\textbf{0.7445}} & {\textbf{0.7059}} & {\textbf{0.7259}} & {\textbf{0.6521}} & {\textbf{0.6572}} & {\textbf{0.6703}} & {\textbf{0.5898}} \\
\bottomrule[1.5pt]
\end{tabular}
\caption{Comparison with different methods on the XFUN \emph{w.r.t} F1 score ($\uparrow$), where ``SER'' denotes the semantic entity recognition and ``RE'' denotes the relation extraction.}\label{tab:sota}
\end{table*}

\begin{table}[h]
\centering
\resizebox{0.46\textwidth}{!}{
\begin{tabular}{cc|c} 
\toprule[1.5pt]
Methods  & Modality  & SER  \\ \hline \hline
BERT\cite{devlin2018bert}   & Language  & 0.6026\\
RoBERTa\cite{conneau2019unsupervised}  &  Language  & 0.6648 \\
BROS\cite{hong2021bros}  &  Language & 0.8121\\
LayoutLMv1\cite{xu2020layoutlm} &   Language + Layout+ Vision    &0.7927 \\
LayoutXLM\cite{xu2021layoutxlm}  &  Language + Layout+ Vision  &0.8276\\
DocFormer\cite{appalaraju2021docformer}  &  Language + Layout+ Vision   & 0.8334\\
SelfDoc\cite{li2021selfdoc} &Language + Layout+ Vision &0.8336\\ 
StructuralLM*\cite{li2021structurallm}     &  Language + Layout &\textbf{0.8514}\\ 
XYLayoutLM   & Language + Layout+ Vision & 0.8335  \\ 
\bottomrule[1.5pt]
\end{tabular}}
\caption{Comparison with different methods on the FUNSD \emph{w.r.t} F1 score ($\uparrow$). The * means StructuralLM use the LARGE model while others use BASE models.}\label{FUNSD}
\end{table}

However, simply replacing the MLP in LayoutXLM to CPE reduces the performance in document understanding tasks. One reason is the wrong neighbors of input tokens due to the improper reading order. Since CPE conditions on local context with convolutions, wrong neighbors will harm the model performances. Another reason is that CPE is designed specialized for image classification. The input visual tokens of image classification are $16\times 16$ patches, and they can be naturally reshaped to 2D for local context extraction. However, in multimodal tasks, we also have 1D textual tokens in our input. These textual tokens only have 1D relations, so that they can not be reasonably reshaped to 2D.

The first problem is solved with a proper reading order obtained by using our Augmented XY Cut. In this section, we propose Dilated Conditional Position Encoding (DCPE) to tackle the second problem, \emph{i.e.}, how to extract 1D local layouts from texts. As shown in Figure~\ref{overview}(a), our DCPE processes the textual and visual features individually. Specifically, DCPE reshapes the 2D visual features and generates their position embeddings following the CPE. While for textual features, we utilized 1D convolutions to extract 1D local layouts. The encoded embeddings from the texts and images are concentrated as the final output.

Another observation is that multimodal tasks often need larger receptive fields while capturing local layouts. For example, in the sentence ``he is a very handsome boy'', the relation of ``he'' and ``boy'' is essential but can not be successfully captured by standard 1D convolutions due to the small convolution kernel size (\emph{e.g.}, 3). To this end, we adopt dilated convolution~\cite{yu2015multi} to replace standard convolutions, aiming for long-range neighbor information with larger receptive fields. Let $l$ be the dilation rate and the dilated convolution $*_l$ can be formulated as:
\begin{equation}
    (F*_lk)(\mathbf{p}) = \sum\limits_{\mathbf{s}+l\mathbf{t}=\mathbf{p}}F(\mathbf{s})k(\mathbf{t})
\end{equation}
where $F,k$ are the input feature map and filter.
By repeatedly stacking dilated convolutions with different dilation rates $l>1$, the DCPE module will pay more attention to the long-range neighbor information. Besides, the dilated convolutions have the same parameters as standard convolutions given the same kernel size, which means our DCPE will not increase the model complexity. Note that new parameters in DCPE are initialized by Xavier and updated with the whole model during training.


\section{Experiments}

\subsection{Setup}

\begin{table}[h]
\centering
\resizebox{0.46\textwidth}{!}{
\begin{tabular}{c|cc|cc} 
\toprule[1.5pt]
     & \multicolumn{2}{c}{SER}   & \multicolumn{2}{c}{RE}                                                \\ 
\multirow{-2}{*}{Methods}       & ZH     & ES   & ZH     & ES    \\ \hline \hline
LayoutXLM      & 0.8924 & 0.7550 & 0.7073 &0.6896\\
+   CPE      & 0.8776 & 0.7306 &0.7059  &0.6812\\
+   2$\times$CPE      & 0.8819 & 0.7412&0.7082  &0.6820 \\
+   2$\times$DCPE     & 0.8952 & 0.7548&0.7097  & 0.6843\\
+ XY Cut       &0.8903  &0.7562 &0.7281  &0.7175\\
+ Aug XY Cut       & 0.9023 & 0.7570&0.7389  & 0.7213\\
+ Aug XY Cut \& CPE & 0.9037 & 0.7597&0.7401  & 0.7236\\ 
+ Aug XY Cut \& DCPE     & \textbf{0.9176} & \textbf{0.7687}& \textbf{0.7445} &\textbf{0.7259} \\ 
\bottomrule[1.5pt]
\end{tabular}}
\caption{Ablation studies of \textbf{XYLayoutLM} on XFUN (Chinese, English) for SER and RE tasks \emph{w.r.t} F1 score ($\uparrow$). 2$\times$ means we use two convolutional layers in this module.}\label{tab:ablation}
\end{table}

\noindent \textbf{Datasets.} Following LayoutXLM~\cite{xu2021layoutxlm}, we conduct experiments on widely used VRDU datasets FUNSD~\cite{funsd} and XFUN~\cite{xu2021layoutxlm}. FUNSD is a form understanding dataset for scanned documents. It contains 199 annotated forms with 31485 words. XFUN is a benchmark for multilingual Form Understanding by extending the FUNSD to 7 other languages, including Chinese, Japanese, Spanish, French, Italian, German, and Portuguese, with 1393 fully annotated forms. Each language includes 199 forms, where the training set includes 149 forms, and the test set includes 50 forms. These two datasets provide the official OCR annotations (bounding boxes and tokens) as the input. 

\noindent \textbf{Tasks.} We focus on two tasks from VRDU, Semantic Entity Recognition (SER) and Relation Extraction (RE). Specifically, SER assigns each token a semantic label from a set of four predefined categories: question, answer, header, or other. For RE, following \cite{xu2021layoutxlm}, we construct the set of relation candidates by generating all possible pairs of input tokens. We utilize a specific embedding layer for every pair to generate token type embedding as the token relation representation. The representations of head and tail are concatenated and fed into a bi-affine classifier. The F1 score is used as the evaluation metric for both two tasks.

\noindent \textbf{Model variants.} We init the weight of our XYLayoutLM with pretrained LayoutXLM$_{BASE}$. We set hidden size $d=768$ and use a 12-layer 12-head transformer. The visual backbone is ResNeXt101-FPN, and the visual features are from its $P_2$ layer in FPN following LayoutXLM. 

\begin{table}[h]
\centering
\resizebox{0.46\textwidth}{!}{
\begin{tabular}{c|cc|cc} 
\toprule[1.5pt]
     & \multicolumn{2}{c}{SER}   & \multicolumn{2}{c}{RE}                                                \\ 
\multirow{-2}{*}{Methods}       & ZH     & ES   & ZH     & ES    \\ \hline \hline
default order      & 0.8924 & 0.7550 &0.7073 &0.6896 \\
remove-pos-embed     & 0.8842 &0.7477  & 0.6941 &0.6682 \\
descending (Y,X)      &0.8857  & 0.7486 & 0.7297 &0.7179 \\
descending (X,Y)      & 0.8561 & 0.7343 & 0.6858 &0.6549 \\
descending (X+Y)     &0.8844  &0.7513 &0.7235  &0.7086  \\
XY Cut     &0.8903  &0.7562 &0.7281  &0.7175  \\
Aug descending (Y,X)      &0.8925  & 0.7543 & 0.7331 &0.7212 \\
Aug XY Cut ($0.5,0.5,1$)     &0.8913  &0.7568 &0.7282  &0.7178 \\ 
Aug XY Cut ($0.2,0.2,5$)     & 0.9011 &0.7586  &0.7387  &0.7202 \\ 
Aug XY Cut ($0.5,0.5,5$)        & \textbf{0.9023} & \textbf{0.7600} & \textbf{0.7389} & \textbf{0.7213}\\ 
Aug XY Cut ($0.7,0.7,5$)     & 0.8918 &0.7541  &0.7260  &0.7166 \\ 
Aug XY Cut ($0.5,0.5,10$)        & 0.8702 &0.7399  &0.6920 &0.6894 \\ 
\bottomrule[1.5pt]
\end{tabular}}
\caption{F1 scores ($\uparrow$) of baseline \textbf{LayoutXLM} based on different reading orders on XFUN (Chinese, English).}\label{tab:xycut}
\end{table}

\noindent \textbf{Training details.} We use the same hyper-parameters with LayoutXLM for fair comparisons on two datasets. On XFUN, the learning rate and batch size are set as $5\times 10^{-5}$ and 32 for the SER task, respectively. We train the model with 1000 iterations for convergence. For the RE task, the batch size is 8 with 2500 iterations for training. While on the FUNSD, the batch size is 16, and we train the model for 1000 iterations following \cite{xu2020layoutlmv2}. 

\subsection{Main Results}

Here we compare our method with textual-based methods XLM-RoBERT~\cite{conneau2019unsupervised}, InfoXLM~\cite{chi2020infoxlm} and LayoutXLM~\cite{xu2021layoutxlm} on XFUN. The results are shown in Table~\ref{tab:sota}. From the table we can observe that XYLayoutLM achieves the best performance among the listed methods. More specifically, among the multimodal methods, XYLayoutLM outperforms the original LayoutXLM~\cite{xu2021layoutxlm} by \textbf{1.48\%} F1 score on the XFUN dataset for the SER task. Besides, our XYLayoutLM achieves a 0.6779 F1 score in RE task, which is an obvious improvement beyond the baseline LayoutXLM (0.6432). 
Similar conclusions are drawn on the FUNSD dataset as shown in Table~\ref{FUNSD}. 
Our XYLayoutLM achieves comparable performances with latest methods like DocFormer\cite{appalaraju2021docformer} and SelfDoc\cite{li2021selfdoc}. Note that StructuralLM*\cite{li2021structurallm} uses the LARGE model to get the best performance while other methods in this table only use the BASE model.

\begin{table}[t]
\centering
\resizebox{0.46\textwidth}{!}{
\begin{tabular}{cc|cc|cc} 
\toprule[1.5pt]
\multicolumn{2}{c}{DCPE}       & \multicolumn{2}{c}{SER}   & \multicolumn{2}{c}{RE}                                                \\ \hline 
Text & Image & ZH & ES & ZH & ES \\ \hline \hline 
Conv2d & Conv2d &0.9037  & 0.7597 &0.7059  & 0.6812\\ 
Conv1d & Conv1d & 0.9091 &0.7613 &0.7066  &0.6832\\ 
Conv1d & Conv2d & 0.9140 & 0.7625 & 0.7256 &0.7106\\
D-Conv1d & Conv2d &0.9163  & 0.7669 &0.7440  &0.7244\\ 
Conv1d & D-Conv2d & 0.9149 &0.7642 &0.7427  & 0.7211\\ 
D-Conv1d &  D-Conv2d   & \textbf{0.9176} & \textbf{0.7687} & \textbf{0.7445}  &\textbf{0.7259}\\ 
\bottomrule[1.5pt]
\end{tabular}}
\caption{F1 scores ($\uparrow$) of \textbf{XYLayoutLM} based on different DCPE architectures on XFUN (Chinese, English).}\label{tab:dilated}
\end{table}

Another observation is that only using CPE~\cite{chu2021conditional} on the LayoutXLM for position embeddings generation harms the performances on all tasks. This observation verifies our claim on the weakness of CPE for multimodal networks as mentioned in Section \ref{dcpe}.

\subsection{Ablation Studies}


%

We perform SER experiments on the Chinese and English subsets of the XFUN dataset for the ablation studies. First, we show the impact of progressively integrating our different
components: the DCPE and Augmented XY Cut module, to the baseline in Table~\ref{tab:ablation}. Then we explore different settings of each component individually.

\noindent \textbf{Analysis for Components.} As shown in Table~\ref{tab:ablation}, we first use the CPE to generate position embeddings instead of MLP in LayoutXLM, which decreases about 0.2\% F1 score. It can be explained that the local context obtained by CPE for position embeddings generation is noisy due to the unreasonable reading order and wrong neighbors. In the third and fourth rows, we replace the CPE with our proposed DCPE with dilated convolutions, resulting in an improvement on all tasks. Note that $2\times$ means we stack two convolutional layers in each module since only one dilated convolutions may lose the information in the holes.

However, the performance gain of only using DCPE suffers from improper reading orders. Next, we only add XY Cut to the baseline, leading to significant improvements of F1 score on both SER (1\%) and RE (3\%) tasks, which verifies the essential effect of proper reading order. Furthermore, when we only perform Augmented XY Cut (Aug XY for short in table), its improvement is satisfactory. The last two rows show the performances of CPE and DCPE associated with Augmented XY Cut. We can observe that our DCPE achieves much better results than CPE because of the larger receptive fields on both textual and visual features.

Moreover, benefitting from the Augmented XY Cut, the improvements of DCPE are highly promoted. When adapted to the baseline model in the default improper reading order, DCPE only gains 0.1\% F1 score improvement on the Chinese subset of XFUN in the SER task. However, after Augmented XY Cut, the improvement comes to 1.5\%. 

In total, the whole improvements of XYLayoutLM upon baseline LayoutXLM indicates the effectiveness of our two contributions, Augmented XY Cut and DCPE.

\noindent \textbf{Analysis for Augmented XY Cut.} The token reading order is an essential factor of the effective document understanding method. Thus, to evaluate the improvement achieved by our proposed Augmented XY Cut, we conduct experiments on XFUN with different reading orders based on the baseline LayoutXLM as shown in Table~\ref{tab:xycut}. When we removed all the position embeddings as shown in the second row, the performance decreased for SER and RE tasks, which indicates the importance of position embeddings for incorporating the reading order. The next four rows are heuristic rules for sorting tokens, \emph{i.e.}, descending first by Y-axis then X-axis, first by X-axis then Y-axis, by Y$+$X based on the left-top point of the token box and traditional XY Cut. However, their performances are not satisfying compared to the baseline. Finally, by using our Augmented XY Cut, the model achieves the best performance. Note that we set the hyper-parameter $\lambda_x,\lambda_y,\theta$ as $0.5,0.5,5$ since it has slightly better performance than other choices.

\begin{figure*}[h]
\centering
\includegraphics[width=0.8\textwidth]{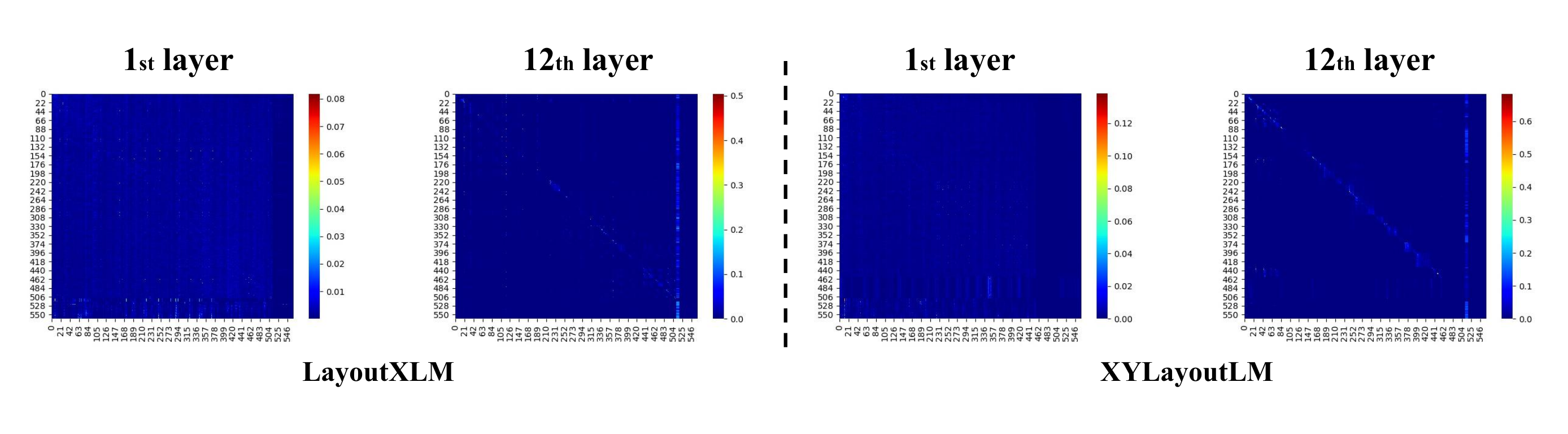}
\caption{Visualizations of the attention scores from one sample based on LayoutXLM and XYLayoutLM. The attention score maps are from the twelveth attention head in the first/twelveth attention layer. Best viewed in color.}
\label{fig4}
\end{figure*}

\begin{figure}[h]
\centering
\includegraphics[width=0.46\textwidth]{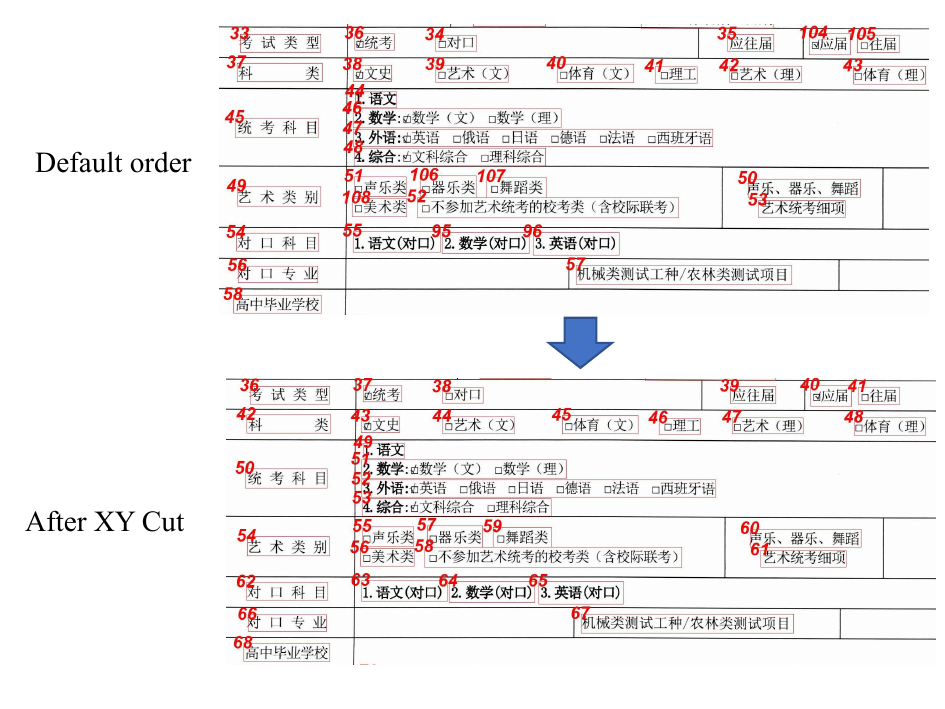}
\caption{Visualizations of Augmented XY Cut on input tokens. The reading order is shown as the red indices.}
\label{fig3}
\end{figure}

\noindent \textbf{Analysis for DCPE.} Besides the ability to deal with various-length inputs, the DCPE module also plays a critical role in our XYLayoutLM for gathering local layouts from both textual and visual features. Thus, we compare several settings inside the DCPE module to improve its effectiveness. As presented in Table~\ref{tab:dilated}, with the replacement of standard convolutional layers to dilated ones for textual and visual tokens, the network performance improves steadily and achieves the peak F1 score when using dilated convolutions for both textual and visual tokens. Another observation is that 1D convolutions can better extract textual features than 2D ones, which also verifies our claim on the reasons for the failure of CPE in multimodal networks.

With these ablation studies, we conclude that in XYLayoutLM: the Augmented XY Cut and DCPE module all play essential roles \emph{w.r.t.} the final performance.

\subsection{Visualizations}


\begin{figure}[h]
\centering
\includegraphics[width=0.46\textwidth]{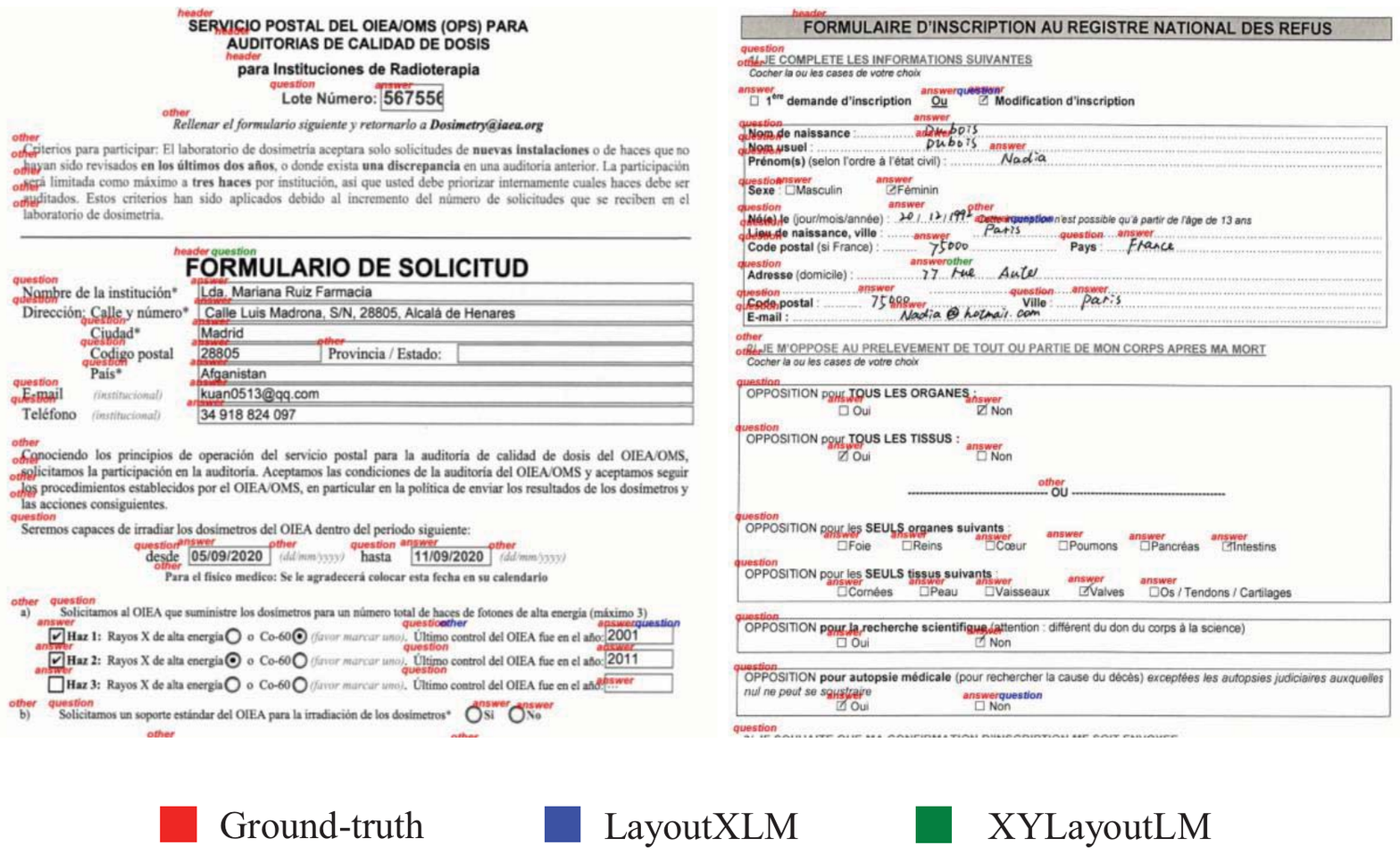}
\caption{Predictions of LayoutXLM and XYLayoutLM for SER task. The red words are the ground truth. The blue and green words are the prediction of LayoutXLM and XYLayoutLM, respectively. Best viewed in color.}
\label{fig2}
\end{figure}

\noindent \textbf{Effects on attention scores.} We have shown that XYLayoutLM can have better performance than the original baseline LayoutXLM. However, because the Augmented XY Cut and DCPE provide the layout information implicitly in the position embeddings, it is interesting to see the attention weights of the transformers. Given a document, the size of the attention score is 561$\times$ 561 following LayoutXLM (512 textual tokens and 49 visual tokens). We visualize the attention score matrix from different attention layers of one sample with the same reading order in Figure~\ref{fig4}. Note that these attention score maps are all from the twelfth attention head without normalization.

From Figure~\ref{fig4} we can draw the following conclusions. The attention weights of XYLayoutLM are larger than LayoutXLM in most layers, which means XYLayoutLM can better capture the attention and relations among tokens. 
Moreover, benefit from DCPE, XYLayoutLM can extract more layout information from the local neighbors since the bright lines in XYLayoutLM are bolder. 

\noindent \textbf{Augmented XY Cut.} We visualize the tokens reading order before and after our proposed Augmented XY Cut in Figure~\ref{fig3}. The figure shows that our XY Cut successfully sorts the input tokens in proper reading order.

\noindent \textbf{Performances on XFUN.} The visualization of XYLayoutLM and LayoutXLM on the XFUN dataset is shown in Figure~\ref{fig2}. The red color in this figure denotes the ground truth category, while the blue and green colors mean the predicted categories of LayoutXLM and XYLayoutLM, respectively. The figure shows that our XYLayoutLM can classify tokens better in challenging situations than LayoutXLM.

\section{Conclusion}
In this work, we introduced XYLayoutLM, a simple yet effective multimodal network for document understanding. Our model contains two related contributions, \emph{i.e.}, Augmented XY Cut for proper reading order and DCPE for generating various-length position embeddings with local layout information. Moreover, it achieves competitive results on several VRDU datasets. We hope our work could inspire designing new frameworks to tackle the challenging document understanding tasks.

\section{Acknowledgements}

The work is supported by Ant Group through Ant Research Intern Program, the Shanghai Municipal Science and Technology Major Project, China (20511100300, 2021SHZDZX0102) and the National Science Foundation of China (62076162).

\section{Appendix}

\subsection{Pseudo-code}
The pseudo-code is shown in Algorithm~\ref{supp_alg}.
\subsection{Inference time for LayoutReader and our method}
We show the inferenece time for our XYLayoutLM and LayoutReader in Table~\ref{app_tab2}. The $*1024$ means it need to decode 1024 times for one document by default.

\begin{algorithm}[t]
\caption{Augmented XY Cut Algorithm}\label{supp_alg}
\begin{algorithmic}[1]
\Require boxes: $B=\{b_i\}_{i=1}^K$, thresholds: $\lambda_x,\lambda_y,\theta$
\Ensure proper reading order: $O = \{s(i)\}_{i=1}^K$
\Function {CUT}{$boxes, n, result, tmp, direction$}  
                \If {$len(boxes) = 0\OR n$}
                    \State \Return{$result$}
                \EndIf
                \State \SORT $boxes$ by $direction$  \Comment{also \SORT $tmp$}
                \If {$direction$ is Y-axis}
                    \State $next \gets$ X-axis
                \ElsIf{$direction$ is X-axis}
                    \State $next \gets$ Y-axis
                \EndIf 
                \State $cur \gets 0$
                \State $sets \gets$ \PROJECT $boxes$ to $direction$
                \For {$i$ in range$(len(boxes))$}
                    \State $set \gets sets[i]$
                    \If {$set\cap sets[i:] = \emptyset$}
                        \State $result \pluseq$ CUT$(boxes[cur:i+1], i-cur, [\ ], tmp[cur:i+1], next)$
                        \State $cur \gets i+1$
                    \EndIf 
                    \EndFor
                \If {$cur \neq i+1$}
                    \State $result \pluseq tmp[cur:i+1]$
                \EndIf  
                \State \Return{$result$}  
                \EndFunction  
\State $tmp \gets$ range$(K)$
\For {$i$ in range$(len(B))$}
    \State $b \gets B[i]$
    \State Random init $v_x\in N(-1,1),v_y\in N(-1,1)$
    \If {$|v_x|>\lambda_x$}
        \State $b[0]\pluseq \theta\cdot v_x, b[2]\pluseq \theta\cdot v_x$
    \EndIf  
    \If {$|v_y|>\lambda_y$}
        \State $b[1]\pluseq \theta\cdot v_y, b[3]\pluseq \theta\cdot v_y$
    \EndIf  
\EndFor
\State $O \gets$ CUT$(B,K,[\ ],tmp,$Y-axis$)$
\end{algorithmic}
\end{algorithm}

\begin{table}
\centering
\begin{tabular}{c|ccc}
\hline
Method          & Inference time/one doc  &Device  \\ \hline
LayoutReader         & (10.3ms ± 34.1us)*1024 &one V100     \\ 
XY Cut      & 8.99ms ± 28.3 µs &CPU  \\ \hline 
\end{tabular}
\caption{Inference time for detecting reading order.}
\label{app_tab2}
\end{table}

{\small
\bibliographystyle{ieee_fullname}
\bibliography{PaperForReview}
}

\end{document}